%% file: main.tex
\def\year{2021}\relax
\documentclass[letterpaper]{article} 
\usepackage{aaai21}  
\usepackage{times}  
\usepackage{helvet} 
\usepackage{courier}  
\usepackage[hyphens]{url}  
\usepackage{graphicx} 
\usepackage{natbib}  
\usepackage{caption} 
\input{math_commands.tex}

\usepackage{soul}
\usepackage{url}
\usepackage[hidelinks]{hyperref}
\usepackage[utf8]{inputenc}
\usepackage{amsmath}
\usepackage{amsthm}
\usepackage{booktabs}
\usepackage{algorithm}
\usepackage{multirow} 
\usepackage{algorithmic}
\usepackage{bbm}
\renewcommand{\defeq}{\overset{\mathrm{def}}{=}}

\renewcommand{\r}{\mathbb{R}}

\newcommand{\y}{\mathcal{Y}}

\title{A Framework of Randomized Selection Based Certified Defenses Against Data Poisoning Attacks}

\title{A Framework of Randomized Selection Based Certified Defenses Against Data Poisoning Attacks}
\author {
        Ruoxin Chen,\textsuperscript{\rm 1}
        Jie Li, \textsuperscript{\rm 1}
        Chentao Wu, \textsuperscript{\rm 1}
        Bin Sheng, \textsuperscript{\rm 1}
        Ping Li, \textsuperscript{\rm 2}\\
}
\affiliations {
    \textsuperscript{\rm 1} Shanghai JiaoTong University\\
    \textsuperscript{\rm 2}The Hong Kong Polytechnic University\\
}
\begin{document}

\maketitle

\begin{abstract}
    Neural network classifiers are vulnerable to data poisoning attacks, as attackers can degrade or even manipulate their predictions thorough poisoning only a few training samples. However, the robustness of heuristic defenses is hard to measure. Random selection based defenses can achieve certified robustness by averaging the classifiers' predictions on the sub-datasets sampled from the training set. This paper proposes a framework of random selection based certified defenses against data poisoning attacks. Specifically, we prove that the random selection schemes that satisfy certain conditions are robust against data poisoning attacks. We also derive the analytical form of the certified radius for the qualified random selection schemes. The certified radius of bagging derived by our framework is tighter than the previous work. Our framework allows users to improve robustness by leveraging prior knowledge about the training set and the poisoning model. Given higher level of prior knowledge, we can achieve higher certified accuracy both theoretically and practically. According to the experiments on three benchmark datasets: MNIST 1/7, MNIST, and CIFAR-10, our method outperforms the state-of-the-art.
    
\end{abstract}
\section{Introduction}
\label{intro}
\input{introduction}
\section{Related Works}
\label{rela}
\input{related}

\section{Preliminaries}
\label{prob}
\input{problem}
\section{Proposed Framework}
\label{meth}
\input{method}

\input{experiments}

\section{Conclusion}
We propose a general framework of random selection based certified defenses against data poisoning attacks in this work. We prove the certified robustness of all the random selection schemes that satisfy our proposed conditions. The certified radius derived from our framework is tighter than previous works. To make our framework more practical, we consider the leverage of prior knowledge about the training set and the attacker’s capability. Given prior knowledge about the training set, our certified radius can be tighter. Given prior experience of the poisoning model, our training algorithm can achieve better certified accuracy practically. Our framework shows the potential of random selection schemes in defending data poisoning. Our study highlights the use of prior knowledge for the improvement of randomized smoothing. We hope our framework would inspire the future work on random selection based defenses.

\bibliography{ref}
\onecolumn
\newpage 
\section*{Appendix}
\label{appendix}

\input{appendix}
\end{document}

%% file: math_commands.tex
\usepackage{amsmath,amsfonts,bm}

\def\eqref#1{equation~\ref{#1}}

\def\ceil#1{\lceil #1 \rceil}
\def\floor#1{\lfloor #1 \rfloor}
\def\1{\bm{1}}

\DeclareMathAlphabet{\mathsfit}{\encodingdefault}{\sfdefault}{m}{sl}
\SetMathAlphabet{\mathsfit}{bold}{\encodingdefault}{\sfdefault}{bx}{n}

\newcommand{\E}{\mathbb{E}}

\usepackage{amsthm}
\newtheorem{theorem}{Theorem}

\newtheorem{corollary}{Corollary}

\usepackage{mathtools}
\newcommand{\defeq}{\vcentcolon=}

%% file: introduction.tex
Neural network classifiers are vulnerable to designed training samples added in the training set or the testing data \cite{biggio2012poisoning,szegedy2014intriguing}. Manipulating $1\%$ of the dataset can cause the target image to be misclassified at a $90\%$ success rate \cite{huang2020metapoison}. Inserting less than $5\%$ poisoned training samples can make the classifier's feature selection almost randomly \cite{xiao2015is}. There exists many kinds of data poisoning attacks \cite{chen2017targeted,liu2018trojaning,turner2018clean,zhao2020clean,turner2019label,ji2017backdoor,yao2019regula,zhang2020backdoor,yao2019latent}. Therefore, it is urgent to develop defenses against data poisoning attacks. Many heuristic defenses have been proposed \cite{wang2019neural,chen2018detecting,chen2019deepinspect,gao2019strip,tran2018spectral,liu2019abs,qiao2019defending,Steinhardt2017CertifiedDF} against data poisoning attacks, but the security level of those defenses is hard to measure. To achieve certified robustness, various certified defenses have been proposed, including randomized smoothing based defenses \cite{wang2020on,rosenfeld2020certified,jia2020intrinsic}, loss based defenses \cite{Steinhardt2017CertifiedDF} and differential privacy based defenses \cite{ma2019data}. Compared with other defenses, randomized smoothing has no limitation on the training algorithms and is able to certify robustness for each testing data. To our knowledge, previous works of randomized smoothing defenses only considered one specific smooth scheme. There is a lack of a framework of randomized smoothing based defenses in defending data poisoning attacks.

In this paper, we propose a framework of random selection based defenses against data poisoning attacks, which is a subset of randomized smoothing. Our framework shows that any random selection scheme satisfying certain conditions is robust to data poisoning attacks. Through our framework, we can derive the certified radius for all qualified random selection schemes, including bagging. We also propose a method to enhance robustness by leveraging $2$ types of prior knowledge: the prior knowledge about the training set and the poisoning model. We divide the poisoning model into $6$ cases and the prior knowledge about the training set into $3$ cases. We can improve the derived certified radius theoretically when the poisoning model is weak. When we have high level of prior knowledge about the training set, we propose 2-phase classification to enhance robustness empirically. In addition, our training algorithms use a weight balance method to solve the problem of low certified accuracy when the size of the selection is small. Our contributions are as follow:

\begin{itemize}
    \item We propose a framework of random selection based defenses against data poisoning attacks. The certified radius derived by our framework is tighter than previous works.
    \item We propose a method to enhance robustness by leveraging prior knowledge about the training set and the poisoning model, which makes our framework more practical in real applications.
    \item We evaluate $3$ random selection based defenses across three classification tasks: MNIST 1/7, MNIST and CIFAR-10. Our certified accuracy outperforms previous works.
\end{itemize}

%% file: related.tex
Previous works have investigated in using randomized smoothing to defend data poisoning attacks. The main challenges in applying randomized smoothing are choosing an appropriate smooth scheme and deriving its certified radius. Rosenfeld et al. \cite{Rosenfeld2019CertifiedRT} proposed the random flip scheme to defend label flipping attacks, but Rosenfeld has not considered the attacks of modifying training data. Wang et al. \cite{wang2020on} proposed the smooth scheme of adding noise into each training sample to defend backdoor attacks, but the certified radius derived by Wang is loose. Jia et al. \cite{jia2020intrinsic} proposed the bagging smooth scheme, which is a variant of random selection schemes. However, Jia has not considered other random selection schemes in defending data poisoning attacks. We also find that the certified radius derived by Jia still has room for improvement. The randomized smoothing defenses also have their intrinsic limitation. Weber and Kumar et al. \cite{weber2020rab,kumar2020curse} have proved the no-go theorem for randomized smoothing independently: \textit{without more information than label statistics under random input perturbations, the largest $\ell_p$ certified radius decreases as $O(d^{1/p-1/2}))$ with dimension $d$ for $p>2$.} This theorem proves the upper bound for the robustness of randomized smoothing methods. To bypass this limitation, leveraging prior knowledge is necessary. Jia et al. \cite{jia2020intrinsic} propose to use transfer learning to improve robustness. However, the improvement of transfer learning methods is defending on the pre-train models. In many cases, a satisfactory pre-train model is not available. Instead of pre-train models, we propose to use prior knowledge about the poisoning model and the training set to improve robustness. The prior knowledge about the training set is the clean training samples known in advance, which is more practical than pre-train models.

%% file: problem.tex
\subsection{Neural Network}
A neural network $h(x \vert D_n):\mathbb{R}^d \rightarrow \mathcal{Y}$ is a mapping from the input feature space to the output space. We refer to the input space as $\mathbb{R}^d$ and the output space as $\mathcal{Y} \defeq \left\{y: y=0,1,\dots,k \right\}$. The confidence for each class is $h^c(x): \mathbb{R}^d \rightarrow \left[0,1\right] (c=1,2,\dots,k)$. We rank each class by its confidence in decreasing order $c_1(x),c_2(x),\dots,c_k(x)$. We denote the confidence of $c_i(x)$ as $p_i(x) \defeq h^{c_i}(x)$. 

\subsection{Classifier}
We refer to $D_n \defeq \left\{(x_i,y_i)_{i=1}^n:x_i\in\r^d, y_i\in \y \}^k \right\}$ as the training set \footnote{We only consider the training sample of unique label}. $h(x \vert D_n)$ is the classifier trained on the dataset $D_n$. In this paper, we mainly focus on the relation between the predictions and the training set. We express $h(x \vert D_n)$ as $f_x(D_n)$, as a function of $D_n$ when $x$ is fixed. $f_x(D_n)$ can be any deterministic or random function.

\subsection{Poisoning model}
Poisoning model is prior knowledge about the attacker's capability. We denote the poisoning model as $P_\rho: D_n \rightarrow \left\{D'_m\right\}$ that maps a dataset to a set of all possible poisoned datasets. $\rho$ is the intensity of the poisoning attacks. In this paper, we consider $6$ cases of the poisoning model: 
\begin{enumerate}
  \item {(\rm $P^1_\rho$)} Insert at most $\rho$ arbitrary samples into $D_n$.
  \item {(\rm $P^2_\rho$)} Delete at most $\rho$ arbitrary samples in $D_n$.
  \item {(\rm $P^3_\rho$)} Modify at most $\rho$ arbitrary samples in $D_n$.
  \item {(\rm $P^4_\rho$)} Insert and/or modify at most $\rho$ arbitrary samples in $D_n$.
  \item {(\rm $P^5_\rho$)} Delete and/or modify at most $\rho$ arbitrary samples in $D_n$.
  \item {(\rm $P^6_\rho$)} Insert, delete and/or modify at most $\rho$ arbitrary samples in $D_n$.
\end{enumerate}
We do not consider the case of \textbf{Insertion and deletion} because the effect of deleting $k$ training sample and inserting $k$ new sample simultaneously is equivalent to modifying $k$ training sample. In the later of this paper, we will show how to leverage the information about the poisoning model to improve robustness.

\subsection{Random selection scheme}
A random selection scheme is a scheme of selecting a subset from the dataset $\mu: D_n \rightarrow D_{sub}$. Here $D_{sub}=\left\{ (x_i,y_i)_{i \in S}: S\subset \left\{1,2,\dots,n \right\}\right\}$. Random selection schemes will not modify any sample in $D_n$.

\subsection{Certified radius}
Given a classifier $f_x()$ and the poisoning model $P_\rho$, the certified radius of $f_x()$ at $x$ under $P_\rho$ is the maximum value of $\rho$ that satisfies:
\begin{align}
  & f_x(D'_m)=f_x(D_n), \; \forall D'_m \in P_\rho(D_n) \nonumber
\end{align}
Certified radius means the maximum attack intensity that the defense can defend.

%% file: method.tex
\begin{figure*}[h]
  \centering
  \includegraphics[width=17cm]{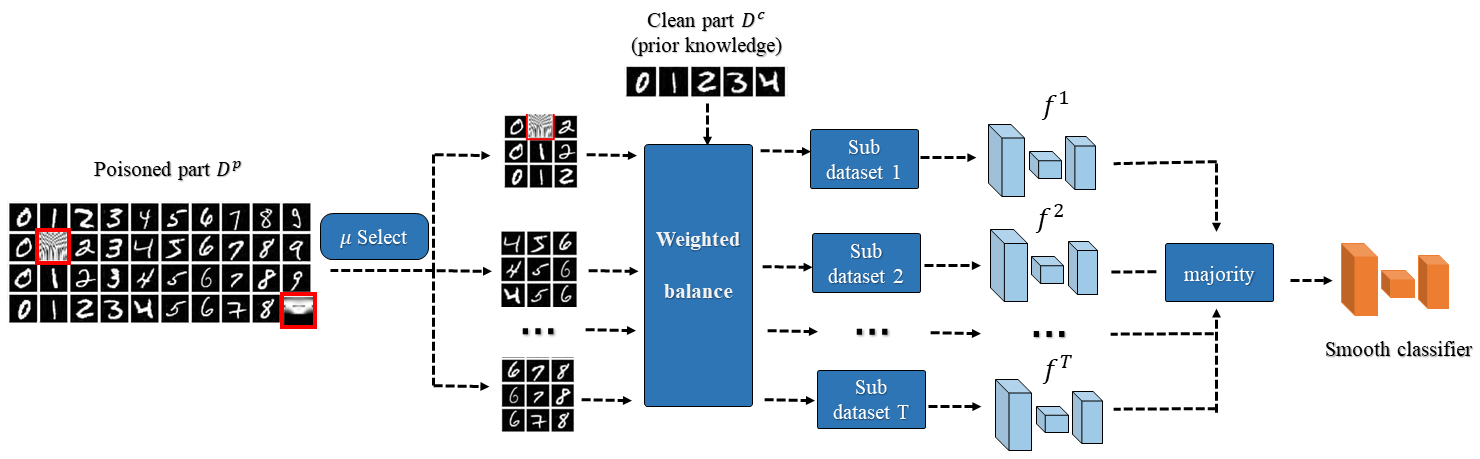}\\
  \caption{A sketch map of random selection based defenses. We sample $T$ sub-datasets $\left\{D^i_{sub}\right\}_{i=1}^T=\mu(D^p)$ from the poisoned part $D^p$ of the training set. $\mu$ is the random selection scheme. We combine $\left\{D^i_{sub}\right\}_{i=1}^T$ with the clean part $D^c$ of the training set to generate the training set for the base classifiers $D_{sub} \cup D^c$. We train $T$ base classifiers on $T$ datasets $\left\{D^i_{sub} \cup D^c\right\}_{i=1}^T$. The smooth classifier $g$ is the aggregation of those $T$ base classifiers by majority rule.} 
  \label{fig-overview}
\end{figure*}

In this paper, our goal is to solve the following $2$ problems:
\begin{enumerate}
  \label{problems}
  \item Which random selection schemes are robust to the poisoning model?
  \item How to compute the certified radius of the random selection schemes that are robust to the poisoning model?
\end{enumerate}
In this section, we solve the above 2 problems by Theorem \ref{theorem-overall}. To show the applicability of Theorem \ref{theorem-overall}, we propose $3$ commonly used variants of random selection schemes. All $3$ schemes satisfy the conditions proposed in theorem \ref{theorem-overall}. Finally we propose $2$ training algorithms for different cases of prior knowledge about the training set to apply random selection schemes into practice.

To achieve certified robustness against the attack where training samples are arbitrarily changed, we first assume that the prediction will change if the training set contains greater than or equal to one poisoned training sample. The assumption is conservative but reasonable because there is no limitation on the neural networks. For a simple model like logistic regression, changing only one training sample can largely affect its predictions. We propose random selection methods to prevent predictions from being affected by limited number of poisoned training samples greatly. The main idea of random selection is to sample a subset of the training set and learns the base classifiers on the subset, which is showed in Figure \ref{fig-overview}. Theorem \ref{theorem-overall} gives the theoretical guarantee for the robustness of random selection schemes. The proof of theorem \ref{theorem-overall} is in the appendix due to space limitation.
\begin{theorem}
  \label{theorem-overall}
  Given any classifier $f_x()$, the smooth classifier of the random selection scheme $\mu$ is $g_x(\cdot)=\E_{D_{n_s} \sim \mu_{n_s}(D_n)} f_x(\cdot)$. Let $g_x$ predicts top-2 classes $c_1,c_2$ with the confidence $p_1,p_2 \; ( p_1>p_2 )$ respectively. Let $P_\rho: D_n \rightarrow D'_m$ as the poisoning model. Let $\Omega=D_n \cap D'_m$ be the set of unmodified samples. If the random selection scheme $\mu$ satisfies 
\begin{align}
    &\forall D'_m \in P_\rho(D_n),\; \exists \text{ constants } \pi_1, \pi_2 \\
    &\text{ s.t. } \forall D_{sub} \in \Omega\\
    &\pi_2 Pr(\mu(D_n)=D_{sub})\geq  Pr(\mu(D'_m)=D_{sub})\\
    &Pr(\mu(D'_m)=D_{sub}) \geq \pi_1 Pr(\mu(D_n)=D_{sub})
\end{align}
and if $p_1,p_2$ satisfy:
\begin{equation}
  \begin{aligned}
    &\pi_1 p_1-\pi_1 \sum_{D_{sub} \not\subset \Omega} Pr(\mu(D_n)=D_{sub}) \geq \\
    &\pi_2 p_2+\sum_{D_{sub} \not\subset \Omega} Pr(\mu(D'_m)=D_{sub}) \label{inequatlity}
  \end{aligned}
\end{equation} 
Then:\\
  \begin{equation}
    g_x(D'_{m})=c_1
  \end{equation}
\end{theorem}

According to Theorem \ref{theorem-overall}, if we can find $\pi_1,\pi_2$ for the random selection scheme $\mu$, $\mu$ is robust to data poisoning attacks. Given the poisoning model $P_\rho$, we can compute the tight certified radius according to the equation \ref{inequatlity}. To show the applicability of Theorem \ref{theorem-overall}, we give $3$ typical random selection schemes and prove that they all satisfy the conditions.

\subsection{Certified robustness of $3$ random selection schemes}
 We introduce $3$ typical random selection schemes: bagging without replacement, bagging with replacement and binomial selection. Those $3$ schemes all have the property $\pi_1=\pi_2$, so the inequality \ref{inequatlity} can be simplified to be $ p_1-p_2 \geq \delta(\rho)$ where $\delta(\rho)$ is
 \begin{align}
  \sum_{D_{sub} \not\subset \Omega} Pr(\mu(D_n)=D_{sub}) + \frac{1}{\pi_1} \sum_{D_{sub} \not\subset \Omega} Pr(\mu(D'_m)=D_{sub}) \nonumber
 \end{align}
$\delta(\rho)$ is the minimum value of $p_1-p_2$ for $x$ to be robust against poisoning $\rho$ training samples. $\delta(\rho)$ increases in proportion to $\rho$. $\delta(\rho)=0$ when $\rho=0$. We will give analytical forms of $\delta(\rho)$ for $3$ schemes.
\begin{enumerate}
  \item {(\rm Bagging without replacement).} Given $D_n$ and let $n_s$ be the size of $D_{sub}$, bagging without replacement $\mu$ is: 
  \begin{align}
    Pr(\mu(D_n)=D_{sub})=1/{n\choose{n_s}}
  \end{align}
  \item {(\rm Bagging with replacement).} Given $D_n$ and $n_s$, bagging with replacement $\mu$ is. 
  \begin{align}
    Pr(\mu(D_n)=D_{sub})=(\frac{1}{n})^{n_s}
  \end{align}
  \item {(\rm Binomial selection).} For each sample $(x_i,y_i)_{i=1}^n \in D_n$, the probability of $(x_i,y_i)$ being selected is $p=n_s/n$.  
  \begin{align}
    Pr(\mu(D_n)=D_{sub})=p^{n_s}(1-p)^{n-n_s}
  \end{align}
\end{enumerate}
$\delta(\rho)$ of $3$ random selection schemes against the poisoning model $P_\rho^6$ are as follow:
\begin{corollary}
  \label{rs without replacement}
  {\rm(Certified robustness of bagging without replacement).} $\pi_1,\pi_2$ of bagging without replacement is\\
  \begin{align}
    \pi_1=\pi_2={{m}\choose{n_s}}/{{n}\choose{n_s}}
  \end{align}
  $\delta(\rho)$ of bagging without replacement is \\
  \begin{align}
    &\max_{m \in \left[n-\rho,n+\rho\right]} 1-{{max(m,n)-\rho}\choose{n_s}}/{{n}\choose{n_s}}+{{n}\choose{n_s}}/{{m}\choose{n_s}} \nonumber \\ 
    &-{{n}\choose{n_s}}{{max(m,n)-\rho}\choose{n_s}}/{{m}\choose{n_s}}^2  \nonumber
\end{align}
\end{corollary}

\begin{corollary}
    {\rm(Certified robustness of bagging with replacement).} 
    $\pi_1,\pi_2$ of bagging with replacement is\\
    \begin{align}
      \pi_1=\pi_2=(\frac{n}{m})^{n_s}
    \end{align}
    $\delta(\rho)$ of bagging without replacement is\\
    \label{rs with replacement}
    \begin{align}
      &\max_{m \in \left[n-\rho,n+\rho\right]} 1+\left(\frac{m}{n} \right)^{n_s}-2 \left( \frac{max(m,n)-rho}{n} \right)^{n_s}  \nonumber
  \end{align}
\end{corollary}

\begin{corollary}
   {\rm(Certified robustness of binomial selection).} 
   $\pi_1,\pi_2$ of binomial selection is\\
    \begin{align}
      \pi_1=\pi_2=(1-p)^{m-n}
    \end{align}
   $\delta(\rho)$ of binomial selection is\\
   \label{rs binomial}
   \begin{align}
    &\max_{m \in \left[n-\rho,n+\rho\right]} 1+(1-p)^{n-m} \nonumber\\
    &-2(1-p)^{min(n-m,0)+\rho} \; (p=n_s/n)\nonumber
  \end{align}
\end{corollary}

\paragraph{Comparison with previous works} 
To our knowledge, only bagging with replacement \cite{jia2020intrinsic} and differentially-private learners \cite{ma2019data} achieve certified robustness against general data poisoning attacks. The assumption of differentially-private learners is different from us, so it is hard to compare with it theoretically. Therefore, we only compare our theory with \cite{jia2020intrinsic}. Compared with \cite{jia2020intrinsic}, $\delta$ derived by Jia contains $2$ additional non-negative terms $\underline{p_1} -(\floor{\underline{p_1} \cdot n^k}/n^k)$ and $\overline{p_2} -(\ceil{\overline{p_2} \cdot n^k}/n^k)$. In almost all cases, those $2$ terms are greater than $0$. Therefore, the certified robustness derived by us is tighter than Jia.

\paragraph{Prior knowledge about poisoning model $P$}
If we know the poisoning model $P$ in advance, we can add additional restriction \ref{formulate ability} on the variable $m$ and $\delta(\rho)$ to derive lower $\delta(\rho)$. Here we formulate $6$ cases of poisoning model into the form that can be used into our theorem \ref{theorem-overall}.
\begin{enumerate}
  \label{formulate ability}
  \item {(\rm $P_1$)} 
  \begin{align}
    m \in \left[n,n+\rho\right], \sum_{D_{sub} \not\subset \Omega} Pr(\mu(D_n)=D_{sub})=0 \nonumber
  \end{align}
  \item {(\rm $P_2$)} 
  \begin{align}
    m \in \left[n-\rho,n\right], \sum_{D_{sub} \not\subset \Omega} Pr(\mu(D'_m)=D_{sub})=1 \nonumber
  \end{align}
  \item {(\rm $P_3$)} $m=n$
  \item {(\rm $P_4$)} $m \in \left[n,n+\rho\right]$
  \item {(\rm $P_5$)} $m \in \left[n-\rho,n\right]$
  \item {(\rm $P_6$)} $m \in \left[n-\rho,n+\rho\right]$
\end{enumerate} 

\subsection{A comparison of $3$ random selection schemes}
\begin{figure}[h]
  \centering
  \includegraphics[width=8cm]{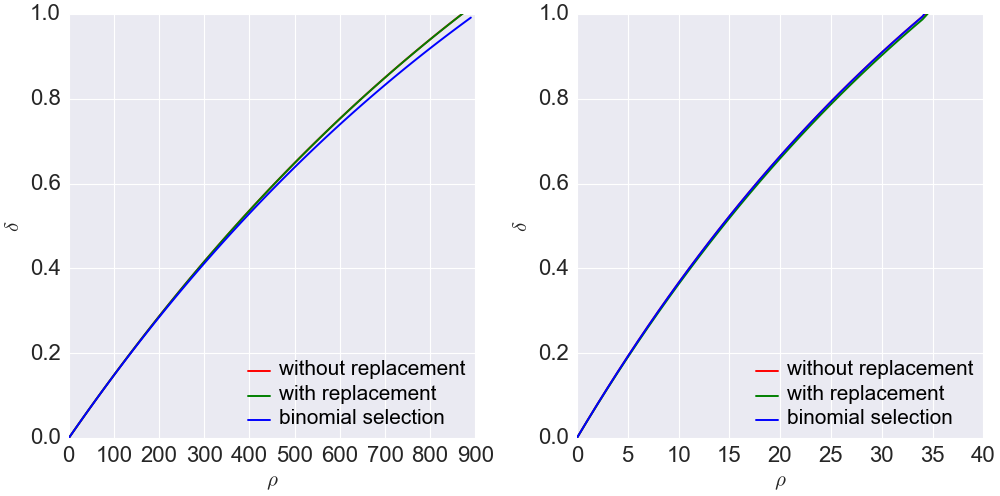}\\
  \caption{$\delta(\rho)$ of $3$ random selection schemes as a function of $\rho$ under the poisoning scheme $P_6$. \textbf{Left:} $n=13007, n_s=10$. \textbf{Right:} $n=50000, n_s=1000$.} 
  \label{fig-comparsion}
\end{figure}
In Figure \ref{fig-comparsion}, we show $\delta(\rho)$ of $3$ random selection schemes respectively. When $n$ and $n_s$ are small, $\delta(\rho)$ of binomial selection is the smallest and $\delta(\rho)$ of bagging without replacement is the highest. When $n$ and $n_s$ are large, $\delta(\rho)$ of $3$ random selection schemes are nearly the same.
 
\subsection{Practical Algorithm}
In practice, evaluating the precise value of $p_1,p_2$ needs unaffordable computation cost. Therefore, we use a Monte-Carlo method to estimate $\underline{p_1},\overline{p_2}$. We follow ~\cite{jia2020certified1} to use SimuEM for the confidence interval estimation. We estimate $\underline{p_1},\overline{p_2}$ with confidence level of $1-\alpha/2$. Then the simultaneous confidence level of $\underline{p_1},\overline{p_2}$ is higher than $1-\alpha$. 
\begin{align}
  &\underline{p_1}=\text{BinoCP}(\alpha/2;T,count(c_1))\\
  &\overline{p_2}= \min(1-\underline{p_1}, \text{BinoCP}(1-\alpha/2;T,count(c_2)))
\end{align}
Our certified prediction \ref{alg-predict} is as follow: 

\begin{algorithm}[!htbp]
  \caption{Certified prediction}
  \label{alg-predict}
  \begin{algorithmic}[1]
  \REQUIRE The input $x$, the smooth classifier $g$, the number of selections $N_s$, the size of poisoned part $N_p$;
  \ENSURE Prediction $y$ and certified radius $r(x)$;
  \STATE Compute $f_i(x)_{i=1}^T$ and counts the number of labels;
  \STATE $c_1,c_2 \leftarrow $ top-$2$ classes in $\left\{f_i(x)\right\}_{i=1}^T$;
  \STATE $count_1,count_2 \leftarrow Count(c_1), Count(c_2)$;
  \STATE $\underline{p_1},\leftarrow \textsc{BinoCP}(\alpha/2,count_1,T)$,\\$\overline{p_2}\leftarrow \textsc{BinoCP}(1-\alpha/2,count_2,T)$;
  \STATE Compute $r(x)=\max \rho$ that satisfies $\delta(\rho) \leq \underline{p_1}-\overline{p_2}$
  \IF{$r(x)$ not exists}
  \RETURN ABSTAIN,ABSTAIN
  \ELSE
  \RETURN $y=c_1$ and $r(x)$;
  \ENDIF
  \end{algorithmic}
  \end{algorithm}

\subsection{Training algorithms and prior knowledge about $D_n$}
\begin{figure}[h]
  \centering
  \includegraphics[width=8cm]{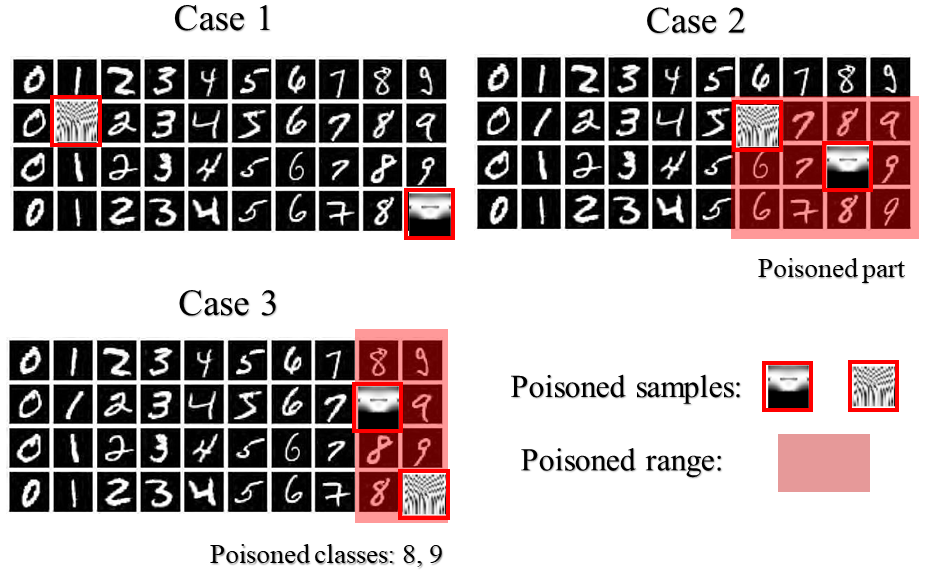}\\
  \caption{A toy example of $3$ conditions. We assume $D_n$ is those $40$ training samples and the attacker can poison at most $2$ samples. \textbf{Case 1}: the attacker modifies $2$ arbitrary samples. \textbf{Case 2}: the attacker poisons $2$ samples in the poisoned part $D^p$. \textbf{Case 3}: the attacker poisons $2$ samples of classes $8,9$.} 
  \label{fig-scenario}
\end{figure}

In reality, users may know the clean part $D^c$ of $D_n$ in advance. In this case, $D_n$ can divided into $2$ parts: the clean part $D^c$ and the potentially poisoned part $D^p$. $D^c=\emptyset$ means that all the training samples have the probability of being poisoned. Treating $D^c$ in the same way as $D^p$ is not ideal. We regard $D^c$ as prior knowledge and only apply random selection schemes on $D^p$. However, how to leverage $D^c$ efficiently is a problem. We divide prior knowledge into $3$ cases as Figure\ref{fig-scenario}.
\begin{enumerate}
  \item {\rm(Case 1).}The attacker can poison all the training samples in $D_n$. ($D^c=\emptyset$)
  \item {\rm(Case 2).}The attacker can operate limited number of training samples in the poisoned part of the training set. ($D^c$ is a set of several training samples)
  \item {\rm(Case 3).}The attacker can operate limited number of training samples of specific classes. ($D^c$ is all the training samples from the unpoisoned classes)
\end{enumerate}

In this section, we propose $2$ training algorithms corresponding to the above $3$ cases. 
\paragraph{Training algorithm for cases 1,2} 
 We sample $T$ sub-datasets of size $n_s$ from $D_n$ and train $T$ base classifiers on those $T$ sub-datasets. In practice, we found that the sampled sub-datasets are usually imbalanced. Especially when $n_s$ is small, the problem of imbalance is highly severe. The base classifiers tend to predict the most frequent class in the sub-datasets. The most frequent class of each sampled sub-dataset has a high probability to be different, which causes $T$ base classifiers tend to make different predictions. Therefore, imbalance not only decreases the accuracy of base classifiers, but also lower the robustness of the smooth classifiers. We use the weighted balance method to eliminate the negative influence brought from the imbalance in $D_n$. The training algorithm for cases 1,2 is showed in Algorithm \ref{alg-c1}

\begin{algorithm}[h]
  \caption{The training algorithm for cases 1,2}
  \label{alg-c1}
  \begin{algorithmic}[1]
  \REQUIRE The potentially poisoned part of the training set $D^p$, the clean part of the training set $D^c$, selection size $n_s$, the number of base classifiers $T$;
  \ENSURE smooth classifier $g$;
  \FOR {$i$ in range($T$)}
  \STATE Sample a $n_s$-size sub-dataset from $D^p$: $D_{n_s}^i=\mu(D^p)$;
  \STATE  Sample $minibatchs$ from $\textsc{BALANCE}(D_{sub}^i \cup D^c)$\\
  $\#$ \textsc{BALANCE}() \textit{is a sample balancing method};
  \STATE  Train $f^i$ on the $minibatchs$;
  \ENDFOR
  \RETURN $g=Maj(f^1,f^2,\dots,f^T)$ $\#$ Majority rule; 
  \end{algorithmic}
  \end{algorithm}

\begin{algorithm}[!htbp]
  \caption{$2$-phase weighted training for case 3}
  \label{alg-c3}
  \begin{algorithmic}[1]
  \REQUIRE The potentially poisoned part of the training set $D^p$, the clean part of the training set $D^c$, selection size $n_s$, the number of base classifiers $T$;
  \ENSURE smooth classifier $g$;
  \FOR {$i$ in range($T$)}
  \STATE Sample a $n_s$-size sub-dataset from $D^p$: $D_{n_s}^i=\mu(D^p)$;
  \STATE Construct training set $D_{sub}^i=D_{n_s}^i \cup D^c$;\\
  $\#$ \textit{All the samples in $D^c$ is of the class $\tilde{c}$}
  \STATE $\tilde{D}_{sub}^i,G_{D_{sub}^i}= \textsc{DETERMINE}(D_{sub}^i,\textsc{HASH()},G)$;
  \STATE  Sample $minibatchs$ from $\textsc{BALANCE}(D_{sub}^i \cup D^c)$\\
  $\#$ \textsc{BALANCE}() \textit{is a sample balancing method}
  \STATE  Train $f^i$ on the $minibatchs$;
  \ENDFOR
  \STATE Train $\phi_2$ on the dataset $D^c$ for the $2$nd phase classification;
  \RETURN $g=\phi_2(Maj(\phi_1^1,\phi_1^2,\dots,\phi_1^t))$ $\#$ Majority rule; 
  \end{algorithmic}
  \end{algorithm}

\begin{figure*}[htbp]
  \centering
  \includegraphics[width=17.5cm]{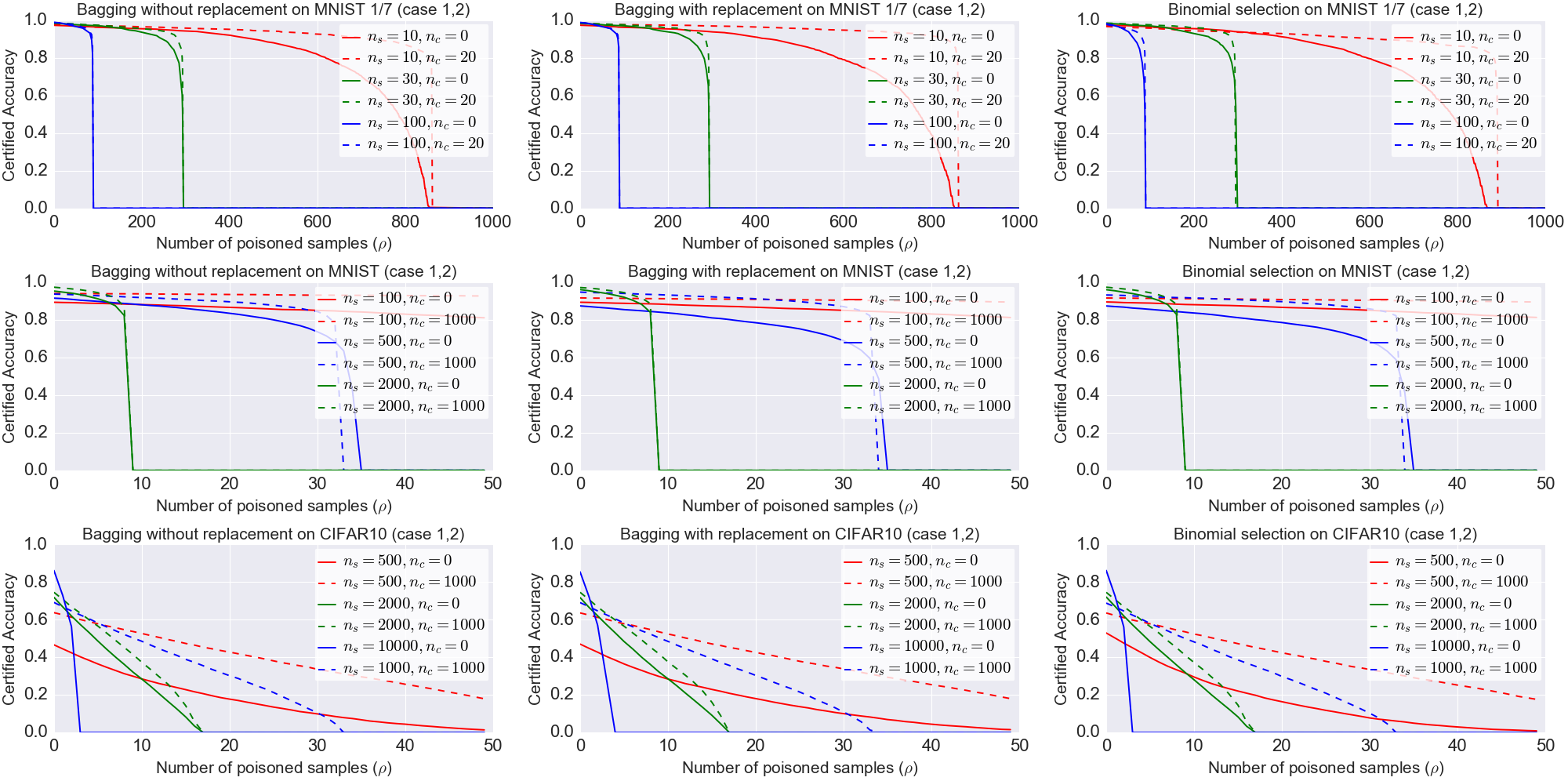}\\
  \caption{Experiments of $3$ random selection schemes on MNIST 1/7 (\textbf{Top row}), full MNIST (\textbf{Middle row}) and CIFAR10 (\textbf{Bottom row}). \textbf{Left}: bagging without replacement \ref{rs without replacement},  \textbf{Middle}: bagging with replacement \ref{rs with replacement} \textbf{Right}: binomial selection  \ref{rs binomial} ($n_s$: the size of the random selection, $n_c$: the number of known clean samples).} 
  \label{exper}
  \end{figure*}

\paragraph{Training algorithm for case 3}
   We refer to the set of clean classes and the set of potentially poisoned classes as $C_{clean},C_{poisoned}$ respectively. $C_{clean} \cup C_{poisoned}$ is the set of all classes. For convenience, we use a virtual class $\tilde{c}$ to represent $C_{clean}$. The main idea is to use $2$-phase classifiers as the base classifiers for case $3$. In the $1$st phase, we use the sub-classifier $\phi_1$ to classify $x$ to be which class among $\tilde{c} \cup C_{poisoned}$. If $\phi_1(x)=\tilde{c}$, we use the sub-classifier $\phi_2$ to classify $x$ to be which class in $C_{clean}$. The base classifier is $\phi_2(\phi_1(\cdot))$. For the $i$th base classifier, the sub-classifier $\phi^i_1$ is trained on the sub-dataset $D^i_{sub} \cup D^c, \;D_{sub} \in D^p$. $\phi_2$ is trained on the clean dataset $D^c$, so $\phi_2$ would not be affected by poisoning attacks. We let all the base classifiers share the same $\phi_2$. If two sub-classifiers $\phi^i_1, \phi^j_1$ predict the same class, the prediction of the base classifiers $\phi^i_1(\phi_2)$ and $\phi^j_1(\phi_2)$ will be the same. Therefore, training $T$ base classifiers is equal to training $T$ sub-classifiers $\left\{\phi^i_1\right\}_{i=1}^T$ on $\left\{D^i_{sub} \cup D^c \right\}_{i=1}^T$ and one $\phi_2$. The original $\# C_{clean}+\# C_{poisoned}$ classification task is converted into a $1+\# C_{poisoned}$ classification task. The complexity of neural networks $\phi_1$ and $\phi_2$ can be much lower than the original base classifiers, which decreases computation cost. 

%% file: experiments.tex

\section{Experiment}

\subsection{Experimental Setup}
To compare fairly, we conduct experiments on three typical datasets: MNIST 1/7 ($13,007$ training samples), MNIST ($60,000$ training samples) and CIFAR10 ($60,000$ training samples), which are the same as previous works. We use LeNet for MNIST 1/7, MNIST and use ResNet 18 for CIFAR10. We train $T=1000$ base classifiers for MNIST, 1/7 MNIST and $T=500$ base classifiers for CIFAR-10. For all base classifiers, we use the weight balance method to train. We found that when the size of $D_{sub}$ is small, the I/O cost of loading $D_{sub}$ is much higher than computation. When $n_s=10$ for MNIST 1/7, the GPU-Util is only $1\%$. Therefore, for each sampled sub-dataset $D_{sub}$, we apply \textbf{WeightedRandomSampler} to expand $D_{sub}$ to a fixed-size dataset. We can achieve convergence after training only a few epochs on the expanded dataset, which vastly accelerates the training. For conciseness, we use scheme 1, 2, 3 to represent the bagging without replacement, the bagging with replacement and the binomial selection respectively.

We evaluate robustness of the defenses by their certified accuracy. Given $\rho$, the certified accuracy of $\rho$ is:
\begin{align}
  CA=\frac{1}{n} \sum_{i=1}^n 1 \left\{g_{x_i}(D_n)=y_i \text{ and } \underline{p_1}-\overline{p_2} \geq \delta(\rho) \right\}
\end{align}
where $\underline{p_1},\overline{p_2}$ is estimated by SimuEM with confidence level $0.999$, which is the same as \cite{jia2020intrinsic}.

\subsection{A comparison on $3$ random selection schemes (case 1, 2)}
\begin{figure}[htbp]
  \centering
  \includegraphics[width=7cm]{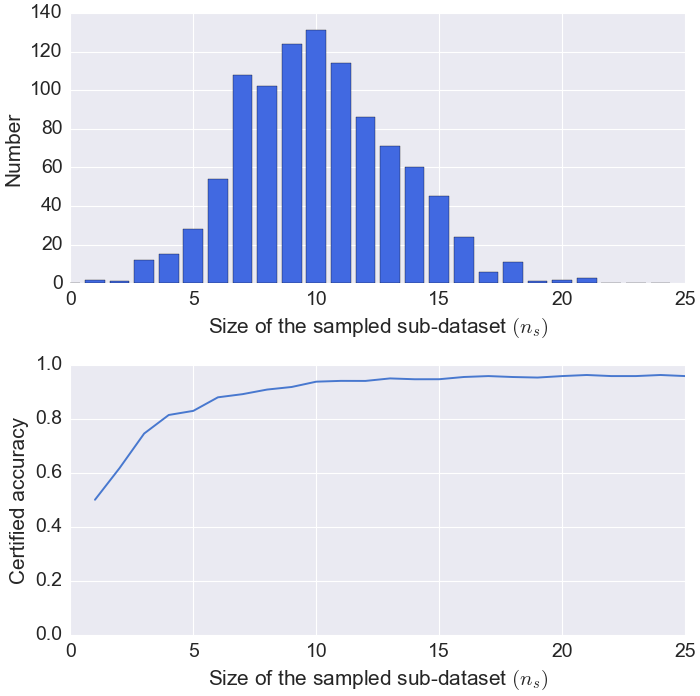}\\
  \caption{\textbf{Top}: the number of the $n_s$-size sub-datasets sampled by the binomial selection. The number of all sub-datasets is $13007$, which is the size of MNIST 1/7 training set. \textbf{Bottom}: the average clean accuracy as a function of the size of the datasets.} 
  \label{scheme3}
  \end{figure}

Figure \ref{exper} shows the performances of $3$ random selection schemes on MNIST 1/7, full MNIST and CIFAR-10. The certified accuracy of $3$ random selection schemes is nearly the same across all the combinations of $n_s$ and $n_c$. However, for the dataset MNIST 1/7 when $n_s=10,n_c=0$, the certified accuracy of scheme 3 decreases to $0\%$ at $868$ while scheme1,2 are at $855,857$ respectively. This is because $\delta(\rho)$ of scheme 3 is less than the $\delta(\rho)$ of scheme 1,2 when the size of the training set $n$ and the selection size $n_s$ are small, which is showed in Figure \ref{fig-comparsion}. Some people may worry that the variation in the sizes of sub-datasets sampled by scheme 3 will cause several base classifiers to perform not well. Figure \ref{scheme3} shows the distribution of the sizes of the sub-datasets and the average accuracy corresponding to the size of training set. Among $T=1000$ sub-datasets, more than $994$ datasets are of the sizes larger than $3$. The accuracy of $994$ base classifiers is higher than $80\%$. MNIST 1/7 is a simple classification task, so the classifiers trained with few samples can achieve high accuracy. On full MNIST and CIFAR-10, $3$ schemes perform almost the same. One reason is showed in Figure \ref{fig-comparsion}: when $n_s+n_c$ is large, $\delta(\rho)$ of $3$ schemes are nearly the same. Another reason is that when the classification task is not simple, the improvement in accuracy needs bigger changes of the dataset. Therefore, the accuracy of $3$ schemes is close.

\subsection{A comparison on case 3}
\begin{figure}[htbp]
  \centering
  \includegraphics[width=8cm]{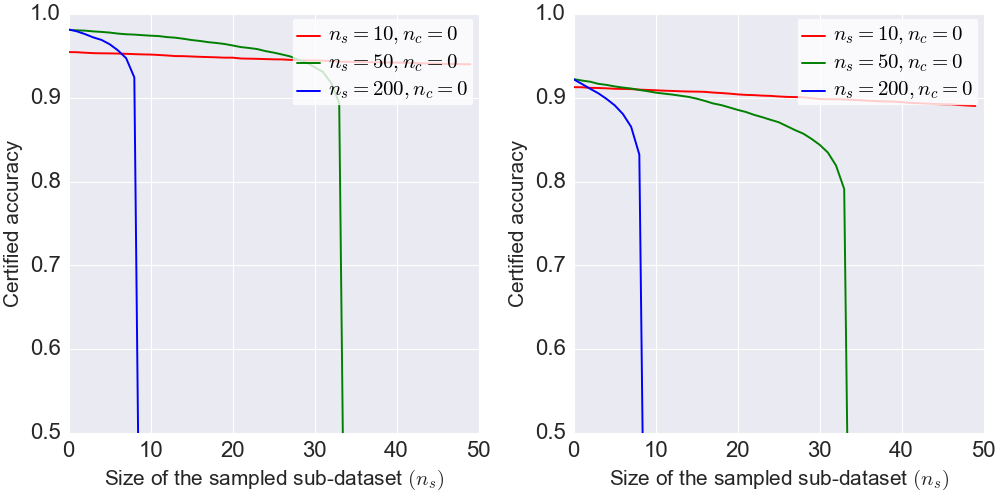}\\
  \caption{\textbf{Left}: the certified accuracy of 2-phase classifiers \ref{alg-c3} as a function of the size of the sub-datasets $n_s$. \textbf{Right}: the certified accuracy of classifiers trained by the algorithm \ref{alg-c1} as a function of the size of the sub-datasets $n_s$.} 
  \label{mnist-c3}
  \end{figure}

We conduct experiments on MNIST to compare the $2$-phase classifiers with the traditional training algorithm used in \cite{jia2020intrinsic}. For a fair comparison, we use the random selection scheme 2, which is the same as \cite{jia2020intrinsic}. Let $0,1,2,3,4,5,6,7,8$ to be the clean classes and $9$ to be the poisoned class. In our $2$-phase classification, the dataset for training the $2$nd-phase sub-classifier $\phi_2$ is well balanced. Therefore, we can easily train $\phi_2$ of the accuracy of classifying $0,1,2,3,4,5,6,7,8$ up to $99.1\%$. For the $1$st-phase sub-classifier $\phi_1$, we use the weight balance method to balance the dataset for the binary classification. In Figure \ref{mnist-c3}, the accuracy of $\phi_1$ is higher than $95 \%$ as $n_s=10$. For all the values of $\rho$, the certified accuracy of the 2-phase classifier outperforms that of the classifier trained by the traditional training algorithm.

\subsection{A comparison with other methods}
We only compare our random selection defenses with bagging method \cite{jia2020intrinsic} because it outperforms other $2$ methods \cite{Rosenfeld2019CertifiedRT,ma2019data} largely in the experiments of \cite{jia2020intrinsic}. The bagging method is the same as scheme 2. For case 1,2, the $\delta(\rho)$ of scheme 2 is larger than scheme 3 that we propose. Therefore, scheme 3 achieve better certified accuracy in MNIST 17 when $n_s=10,n_c=0$ as showed in Figure \ref{exper}. For case 3, we propose 2-phase classifiers, which outperforms previous training algorithms. 

\subsection{Ablation study}
\paragraph{Impact of the random selection scheme}
When the classification task is simple, which means classifiers trained on a small size sub-dataset can already achieve high accuracy, scheme 3 may be the best choice among $3$ random selection schemes. It is because $\delta(\rho)$ of scheme 3 is lower than other $2$ schemes when $n$ and $n_s$ are small. When the classification task is difficult, which means only the classifiers trained on large size sub-datasets can achieve ideal accuracy, $3$ schemes perform nearly the same.

\paragraph{Impact of $n_s$}
$n_s$ is the size of the selection. As $n_s$ increases, each sub-dataset contains more training samples. Therefore, the accuracy of the smooth classifiers increases. Higher $n_s$ also results in higher $\delta(\rho)$, so $\underline{p_1}-\overline{p_2}$ needs to be larger for the same $\rho$. Therefore, $n_s$ controls the trade-off between the clean accuracy and robustness. 

\paragraph{Impact of $n_c$}
$n_c$ is the size of the clean part $D^c$. In all experiments, the certified accuracy at all $\rho$ improves as $n_c$ increases. For the random selection schemes at small $n_s$, the degree of the increase in certified accuracy is higher than that of the high $n_s$.

%% file: appendix.tex
\section{Proof of Theorem 1}
    Let $P_\rho: D_n \rightarrow D'_m$ as the training model on the dataset $D_n$. For all $D'_m \in P_\rho(D_n)$, let $\Omega$ be $D_n \cap D'_m$. If the random selection scheme $\mu$ satisfies 
\begin{align}
    &\forall D'_m \in P_\rho(D_n)\\
    &\exists \text{ constant } \pi_1, \pi_2\\
    &\forall D_{sub} \in \Omega\\
    &\pi_2 Pr(\mu(D_n)=D_{sub})\geq  Pr(\mu(D'_m)=D_{sub}) \geq \pi_1 Pr(\mu(D_n)=D_{sub})
\end{align}

We refer to $g_x(D_n)= \E_{D_{sub}\sim \mu(D_n)} f_x(D_{sub})$ as the smooth classifier ($x$ is arbitrary). Without loss of generality, we assume $g$ predicts $c_1,c_2$ as the top-$2$ classes for $x$. The $g$'s confidence of $c_1,c_2$ is $p_1,p_2$.
\begin{align}
    & \E_{D_{sub}\sim \mu(D_n)} Pr(f_x(D_{sub})=c_1)= p_1\\
    & \E_{D_{sub}\sim \mu(D_n)} Pr(f_x(D_{sub})=c_2)= p_2
\end{align}
Our goal is equal to prove $\forall D'_m \in \left\{P_\rho(D_n)\right\}, \;Pr(g_x(D'_m)=c_1) \geq Pr(g_x(D'_m)=c_2)$\\
First we consider the case that the attacker only modify $D_n$, without insertion or deletion. For the poisoned dataset $D'_m$ that different from $D_n$ at most $\rho$ samples. We denote $S_\rho$ as the set of those $\rho$ poisoned samples. Then we have:
\begin{align}
    & Pr(g_x(D'_m)=c_1)   \\
    =& \E_{D_{sub}\sim \mu(D'_m)} Pr(f_x(D_{sub})=c_1)   \\ 
    =& \sum_{D_{sub} \subset \Omega} Pr(\mu(D'_m)=D_{sub}) Pr(f_x(D_{sub})=c_1 \vert D_{sub})+ \sum_{D_{sub} \not\subset \Omega} Pr(\mu(D'_m)=D_{sub}) Pr(f_x(D_{sub})=c_1 \vert D_{sub})   \\
    \geq & \sum_{D_{sub} \subset \Omega} Pr(\mu(D'_m)=D_{sub}) Pr(f_x(D_{sub})=c_1 \vert D_{sub})+ \sum_{D_{sub} \not\subset \Omega} Pr(\mu(D'_m)=D_{sub}) *0  \label{eq1}  \\
    =&\sum_{D_{sub} \subset \Omega} Pr(\mu(D'_m)=D_{sub}) Pr(f_x(D_{sub})=c_1 \vert D_{sub})   \\
    =&\sum_{D_{sub} \subset \Omega} Pr(\mu(D'_m)=D_{sub}) Pr(f_x(D_{sub})=c_1 \vert D_{sub})
\end{align}

For our assumption about $\mu$, we have: \\
\begin{align}
    &\sum_{D_{sub} \subset \Omega} Pr(\mu(D'_m)=D_{sub}) Pr(f_x(D_{sub})=c_1 \vert D_{sub})   \\
    \geq& \pi_1 \left[  \sum_{D_{sub} \subset \Omega} Pr(\mu(D_n)=D_{sub}) Pr(f_x(D_{sub})=c_1 \vert D_{sub}) \right]   \label{eq2}\\
    =&\pi_1 \left[Pr(g_x(D_n)=c_1)-\sum_{D_{sub} \not\subset \Omega} Pr(\mu(D_n)=\pi D_{sub}) Pr(f_x(D_{sub})=c_1 \vert D_{sub} \right]   \\
    =& \pi_1 \left[ p_1-\sum_{D_{sub} \not\subset \Omega} Pr(\mu(D_n)=D_{sub}) Pr(f_x(D_{sub})=c_1 \vert D_{sub}) \right]   \\
    \geq & \pi_1 \left[ p_1-\sum_{D_{sub} \not\subset \Omega} Pr(\mu(D_n)=D_{sub})*1 \right]   \label{eq3}\\
    =& \pi_1 \left[ p_1-\sum_{D_{sub} \not\subset \Omega} Pr(\mu(D_n)=D_{sub}) \right]
\end{align}
Same as the above proof, we have:
\begin{align}
    &Pr(g_x(D'_m)=c_2)   \\
    =&\sum_{D_{sub} \subset \Omega} Pr(\mu(D'_m)=D_{sub}) Pr(f_x(D_{sub})=c_2 \vert D_{sub})+ \sum_{D_{sub} \not\subset \Omega} Pr(\mu(D'_m)=D_{sub}) Pr(f_x(D_{sub})=c_2 \vert D_{sub})   \\
    \leq & \sum_{D_{sub} \subset \Omega} Pr(\mu(D'_m)=D_{sub}) Pr(f_x(D_{sub})=c_2 \vert D_{sub})+ \sum_{D_{sub} \not\subset \Omega} Pr(\mu(D'_m)=D_{sub}) *1   \label{eq4} \\
    =&\sum_{D_{sub} \subset \Omega} Pr(\mu(D'_m)=D_{sub}) Pr(f_x(D_{sub})=c_2 \vert D_{sub})+\sum_{D_{sub} \not\subset \Omega} Pr(\mu(D'_m)=D_{sub})    \\
    \leq&\pi_2 \sum_{D_{sub} \subset \Omega} Pr(\mu(D_n)=D_{sub}) Pr(f_x(D_{sub})=c_2 \vert D_{sub})+\sum_{D_{sub} \not\subset \Omega} Pr(\mu(D'_m)=D_{sub})   \label{eq5}\\
    =&\pi_2 \left[ Pr(g_x(D_n)=c_2)-\sum_{D_{sub} \not\subset \Omega} Pr(\mu(D_n)=D_{sub}) Pr(f_x(D_{sub})=c_2 \vert D_{sub})  \right]+\sum_{D_{sub} \not\subset \Omega} Pr(\mu(D'_m)=D_{sub})   \\
    =&\pi_2 \left[ p_2-\sum_{D_{sub} \not\subset \Omega} Pr(\mu(D_n)=D_{sub}) Pr(f_x(D_{sub})=c_2 \vert D_{sub}) \right]+\sum_{D_{sub} \not\subset \Omega} Pr(\mu(D'_m)=D_{sub})   \\
    \leq &\pi_2 \left[ p_2-\sum_{D_{sub} \not\subset \Omega} Pr(\mu(D_n)=D_{sub})*0 \right] +\sum_{D_{sub} \subset \Omega} Pr(\mu(D'_m)=D_{sub}) \label{eq6}  \\
    =&\pi_2 p_2+\sum_{D_{sub} \not\subset \Omega} Pr(\mu(D'_m)=D_{sub})
\end{align}
if \\
\begin{align}
    \label{condition}
    \pi_1 p_1-\pi_1 \sum_{D_{sub} \not\subset \Omega} Pr(\mu(D_n)=D_{sub}) \geq \pi_2 p_2+\sum_{D_{sub} \not\subset \Omega} Pr(\mu(D'_m)=D_{sub})
\end{align}
then we have\\
\begin{align}
    Pr(g_x(D'_m)=c_1) \geq Pr(g_x(D'_m)=c_2)
\end{align}
For all the inequalities above \ref{eq1},\ref{eq2},\ref{eq3},\ref{eq4},\ref{eq5},\ref{eq6}, the equalities can hold. Therefore, our bound of the certified radius is tight. The proof of theorem 1 is complete.\\
\subsection{Proof of certified radius bound for $3$ random selection schemes.}
Without loss of generality, we assume $D_{sub}$ is of $n_s$ size. Give $m,n$, the size of $\Omega$ is equal to $max(m,n)-\rho$. For all $3$ random selection schemes that satisfy $\pi_1=\pi_2$, the inequality \ref{condition} can be expressed as 
\begin{align}
    p_1-p_2 \geq \sum_{D_{sub} \not\subset \Omega} Pr(\mu(D_n)=D_{sub}) + \frac{1}{\pi_1} \sum_{D_{sub} \not\subset \Omega} Pr(\mu(D'_m)=D_{sub})
\end{align}
\paragraph{Bagging without replacement}
First we prove $\mu$ satisfies theorem 1
\begin{align}
    &\forall D_{sub} \in \left\{D_{sub}: \; D_{sub} \subset \Omega\right\}\\
    &Pr(\mu(D_n)=D_{sub})= {{n}\choose{n_s}}\\
    &Pr(\mu(D'_m)=D_{sub}) = {{m}\choose{n_s}}\\
    &\pi_1=\pi_2={{m}\choose{n_s}}/{{n}\choose{n_s}}\\
    &\sum_{D_{sub} \not\subset \Omega} Pr(\mu(D_n)=D_{sub})=1-{{max(m,n)-\rho}\choose{n_s}}/{{n}\choose{n_s}}\\
    &\sum_{D_{sub} \not\subset \Omega} Pr(\mu(D'_m)=D_{sub})=1-{{max(m,n)-\rho}\choose{n_s}}/{{m}\choose{n_s}}
\end{align}
The certified radius is $\max \rho$ that satisfies\\
\begin{align}
    \max_{n-\rho \leq m \leq n+\rho} 1-{{max(m,n)-\rho}\choose{n_s}}/{{n}\choose{n_s}} +{{n}\choose{n_s}}/{{m}\choose{n_s}}-{{n}\choose{n_s}}{{max(m,n)-\rho}\choose{n_s}}/{{m}\choose{n_s}}^2  \leq p_1-p_2
\end{align}
\paragraph{Bagging with replacement}
First we prove $\mu$ satisfies theorem 1
\begin{align}
    &\forall D_{sub} \in \left\{D_{sub}: \; D_{sub} \subset \Omega\right\}\\
     &Pr(\mu(D_n)=D_{sub})=(\frac{1}{n})^{n_s}\\
     &Pr(\mu(D'_m)=D_{sub}) = (\frac{1}{m})^{n_s}\\
     &\pi_1=\pi_2=(\frac{n}{m})^{n_s}\\
     &\sum_{D_{sub} \not\subset \Omega} Pr(\mu(D_n)=D_{sub})=1-(\frac{max(m,n)-\rho}{n})^{n_s}\\
    &\sum_{D_{sub} \not\subset \Omega} Pr(\mu(D'_m)=D_{sub})=1-(\frac{max(m,n)-\rho}{m})^{n_s}
\end{align}
The certified radius is $\max \rho$ that satisfies\\
\begin{align}
    \max_{n-\rho \leq m \leq n+\rho} 1+\left(\frac{m}{n} \right)^{n_s}-2 \left( \frac{max(m,n)-\rho}{n} \right)^{n_s} \leq p_1-p_2
\end{align}

\paragraph{Binomial selection $\mu_p$}
First we prove $\mu_p$ satisfies theorem 1
\begin{align}
    &\forall D_{sub} \in \left\{D_{sub}: \; D_{sub} \subset \Omega\right\}\\
     &Pr(\mu(D_n)=D_{sub})=p^{n_s} (1-p)^{n-n_s}\\
     &Pr(\mu(D'_m)=D_{sub})=p^{n_s} (1-p)^{m-n_s}\\
     &\pi_1=\pi_2=(1-p)^{m-n}\\
     &\sum_{D_{sub} \not\subset \Omega} Pr(\mu(D_n)=D_{sub})=1-(1-p)^{min(n-m,0)+\rho}\\
    &\sum_{D_{sub} \not\subset \Omega} Pr(\mu(D'_m)=D_{sub})=1-(1-p)^{min(m-n,0)+\rho}
\end{align}
The certified radius is $\max \rho$ that satisfies\\
\begin{align}
    \max_{n-\rho \leq m \leq n+\rho} 1+(1-p)^{n-m}-2(1-p)^{min(n-m,0)+\rho}\leq p_1-p_2
\end{align}